\documentclass[conference]{IEEEtran}
\IEEEoverridecommandlockouts

\usepackage{cite}
\usepackage{booktabs}       
\usepackage{amsfonts}       
\usepackage{nicefrac}       
\usepackage{microtype}      
\usepackage{xcolor}         
\usepackage{amsmath}
\usepackage{algorithm}
\usepackage{algorithmic}
\usepackage{float}
\usepackage{graphicx}
\usepackage{caption}
\usepackage[caption=false,font=footnotesize]{subfig}

\usepackage{tabularx}
\usepackage{multirow}
\def\BibTeX{{\rm B\kern-.05em{\sc i\kern-.025em b}\kern-.08em
    T\kern-.1667em\lower.7ex\hbox{E}\kern-.125emX}}
\begin{document}

\title{GraphRectify: Graph-Based Transfer of Adversarial Example Detectors Across Neural Networks}

\author{Arash Vashagh and Roozbeh Razavi-Far\\
Trustworthy and Secure AI (TSAI) Lab, Faculty of Computer Science, University of New Brunswick, Canada\\
Email: arash.vashagh@unb.ca, roozbeh.razavi-far@unb.ca}

\maketitle

\begin{abstract}
Adversarial example detectors are often tied to the classifier backbone they were trained on, limiting reuse when the protected model is replaced or upgraded. Directly transferring such detectors across backbones is challenging because different networks generally produce incompatible internal representations. We propose GraphRectify, a graph-based framework for transferring adversarial image detectors across classifier backbones. GraphRectify learns a structured representation of intermediate classifier features and adapts representations from a new backbone to the detector learned on the original model, enabling detector reuse. We evaluate GraphRectify across multiple datasets, backbone architectures, and adversarial attacks, including detector-aware adaptive attacks that jointly target the classifier and detector. Across the complete evaluation matrix, GraphRectify achieves higher aggregate ROC-AUC than training a detector from scratch on the new backbone and the evaluated transfer ablations. The gains are particularly strong for transfers between different backbone families and when sufficient data are available. In contrast, training from scratch remains competitive in the most data-limited settings. These results show that adversarial detection knowledge can transfer effectively across heterogeneous classifier architectures rather than being relearned whenever the protected backbone changes.
\end{abstract}


\section{Introduction}
\label{sec:introduction}

Adversarial example detectors provide an additional test-time mechanism for
identifying inputs that have been deliberately perturbed to mislead a neural
classifier. Existing approaches use signals such as prediction confidence,
input transformations, intermediate feature statistics, and consistency across
internal representations
\cite{hendrycks2016baseline,feinman2017detecting,lee2018simple,yang2020ml}.
Detectors that rely on intermediate representations can exploit information
that is unavailable from the final classifier output alone, but this dependence
also creates a practical limitation. When the protected classifier is replaced
or upgraded, its internal feature space changes, so a detector trained for the
previous model may no longer operate on compatible representations. Retraining
the detector from scratch discards previously learned detection structure and
may require additional labeled data and computation, while directly reusing the
existing detector is unreliable because independently trained or
architecturally different networks do not generally produce aligned internal
representations.

We refer to the original protected classifier as the \emph{source classifier}
and its feature-extracting architecture as the \emph{source backbone}. The new
or replacement classifier is the \emph{target classifier}, and its architecture
is the \emph{target backbone}. The problem we consider in this work is whether the adversarial
detection structure learned for the source backbone can be reused after the protected classifier changes to the target
backbone.

We address this problem with \emph{GraphRectify}, a graph-based framework for
transferring an adversarial example detector across classifier backbones.
GraphRectify converts selected intermediate features into compact
fixed-dimensional nodes and learns relationships among these nodes using a
dynamic graph module. A detector trained using representations from the source
backbone is referred to as the \emph{source detector}. During \emph{source
training}, the source detector learns a graph-level representation that
distinguishes clean and adversarial inputs. After source training, its graph
module and readout are frozen and reused when adapting detection to the target
backbone.

The main difficulty is that simply projecting target-backbone features into
vectors of the same dimension does not ensure that they match the
representation structure expected by the source detector. GraphRectify
therefore performs \emph{target adaptation}, meaning that only the
target-specific interface to the transferred detector is learned for the new
backbone. At each selected stage, a projected representation from the target
backbone forms a \emph{target node}. It is paired with a detached
\emph{source reference node}, obtained from the same input through the frozen
\emph{source branch}, which consists of the source backbone and the
source-side projection modules. A lightweight source-guided rectifier predicts
a bounded residual correction to the target node. The transferred graph module and readout then process the corrected target nodes. In this way,
GraphRectify adapts the target representations to the detection structure
learned from the source backbone without training the target classifier to
reproduce predictions from the source classifier.

This setting differs from conventional knowledge distillation, which typically
transfers predictive behavior from a teacher model to a student
\cite{hinton2015distilling,zagoruyko2016paying}. It also differs from prior work
on detector generalization across attacks, datasets, or models
\cite{9377649,LI2024103791,han2025prediction}, as well as meta-learning methods
that adapt detectors to previously unseen attacks
\cite{ma2019metaadvdet}. Our focus is specifically on reusing a detector whose
inputs depend on the internal representations of one classifier after those
representations change because the protected backbone has been replaced.
GraphRectify separates backbone-specific node construction from the reusable
graph-level detection structure and adapts the interface between them.

We evaluate GraphRectify on CIFAR-10, STL-10, and SVHN across three
architecture transfers:
ResNet18$\rightarrow$ResNet34,
ResNet34$\rightarrow$ResNet18, and
ResNet18$\rightarrow$MobileNetV2.
We use \emph{target data} to denote the labeled data available in the
target-backbone setting for adapting GraphRectify or training a detector
directly on the target backbone. The evaluation considers five target-data
fractions, four detector-aware attacks, and three random seeds. We compare
GraphRectify with \emph{direct transfer}, in which the existing detector is
reused without target adaptation, and with \emph{target-only detectors}, which
are trained from scratch on the target backbone using the same target-data
fraction and without transferring components from the source detector.
GraphRectify improves overall detection performance relative to both baselines.
Its benefit becomes clearer as more target data are available, although
target-only training can remain competitive in extremely data-limited settings.
GraphRectify also improves performance across all three architecture-transfer
settings, including the cross-family
ResNet18$\rightarrow$MobileNetV2 transfer.

The main contributions of this work are as follows:
\begin{enumerate}
    \item We formulate adversarial detector transfer across classifier
    backbones as a representation-adaptation problem in which detection
    structure learned for a source backbone is reused after the protected
    classifier changes to a target backbone.

    \item We propose GraphRectify, which separates backbone-specific node
    construction from a reusable graph-level detector and introduces
    source-guided bounded rectification to make target representations
    compatible with the transferred detection structure.

    \item We evaluate GraphRectify across three datasets, three architecture
    transfers, five target-data fractions, four detector-aware attacks, and
    three random seeds, including target-only baselines, controlled ablations,
    and cross-family transfer to MobileNetV2.
\end{enumerate}

The remainder of this paper is organized as follows.
Section~\ref{sec:background} reviews adversarial example detection, detector
generalization, transfer learning, and graph-based representation modeling.
Section~\ref{sec:proposed_method} presents GraphRectify and its source-training
and target-adaptation procedures. Section~\ref{sec:experimental_setup}
describes the datasets, training configuration, attacks, and evaluation
metrics. Section~\ref{sec:results} presents the experimental results and
ablation studies. Finally, Section~\ref{sec:discussion} discusses the findings,
limitations, and future directions, and Section~\ref{sec:conclusion} concludes
the paper.

\section{Background}
\label{sec:background}

This section reviews the areas most directly related to GraphRectify:
adversarial example detection, transfer and teacher-guided adaptation,
detector reuse across changing settings, and graph-based modeling of internal
representations. We emphasize the distinction between transferring
\emph{classifier behavior} and transferring the \emph{detection structure}
built around a classifier's intermediate representations.

\subsection{Adversarial Examples and Detection}
\label{subsec:adv_detection}

Deep neural networks are vulnerable to adversarial examples, in which small
input perturbations can change a model's prediction while preserving the
semantic content of the original input. Gradient-based attacks such as the
Fast Gradient Sign Method (FGSM) and Projected Gradient Descent (PGD) are
widely used to construct such examples
\cite{Goodfellow2015,madry2017towards,vashaghAttackSurvey,MohammadDefenseSurvey}.

Adversarial example detectors attempt to distinguish these perturbed inputs
from clean samples at inference time. Existing methods use several types of
signals. Confidence-based approaches use statistics derived from the final
classifier output, such as maximum softmax probability
\cite{hendrycks2016baseline}. Transformation-based approaches test whether
predictions remain stable after operations such as bit-depth reduction or
spatial smoothing \cite{xu2017feature}. Other detectors use information from
intermediate representations. Hidden-layer density estimates and dropout
uncertainty have been used to separate clean and adversarial inputs
\cite{feinman2017detecting}, while Mahalanobis-distance methods model
class-conditional feature distributions across network layers
\cite{lee2018simple}.

More recent approaches exploit consistency and changes in internal behavior.
ML-LOO uses feature-attribution statistics and studies generalization across
attack types \cite{yang2020ml}. BEYOND compares an input with augmented
neighbors using self-supervised representations \cite{he2022your}.
Other work detects adversarial inputs through localized changes across
intermediate layers \cite{yun2025few}. Collectively, these approaches show
that intermediate activations contain useful detection signals beyond final
prediction confidence.

A detector that relies on internal representations, however, is naturally
coupled to the backbone that produces those representations. If that backbone
changes, both the dimensionality and distribution of the detector inputs may
change. This dependency is central to the transfer problem studied in this
paper.

Adversarial detectors must also be evaluated against attacks that account for
the detector itself. Adaptive attacks have demonstrated that detectors that
appear reliable against non-adaptive attacks can often be bypassed
\cite{carlini2017adversarial}. AutoAttack further established the importance
of strong and carefully configured adversarial evaluation
\cite{croce2020reliable}. For this reason, our final evaluation uses
detector-aware attacks that jointly target classifier misclassification and
detector evasion.

\subsection{Teacher Guidance and Transfer}
\label{subsec:teacher_transfer}

Teacher--student learning provides mechanisms for transferring information
from a previously trained model to another model. Knowledge distillation
trains a student to reproduce softened outputs of a teacher
\cite{hinton2015distilling,MOSLEMI2024100605}, while attention transfer uses
intermediate teacher representations as additional supervision
\cite{zagoruyko2016paying}. In both cases, the objective is primarily to
transfer predictive knowledge from one model to another.

Transfer has also been studied in adversarial machine learning. Prior work has
examined transfer of adversarial robustness between tasks
\cite{davchev2019empirical}, standalone adversarial detection without access
to the protected model's internal features \cite{moitra2022adversarial},
transferability in audio deepfake detection \cite{farooq2025transferable},
and adversarial transfer based on self-supervised transformer features
\cite{wu2025boosting}. MetaAdvDet uses meta-learning to adapt an adversarial
detector to previously unseen attack types \cite{ma2019metaadvdet}.

These approaches address forms of transfer that differ from the setting
considered here. GraphRectify does not train a target classifier to imitate a
source classifier, nor does it transfer class predictions or robustness from
one classifier to another. Instead, the protected target classifier remains
independently trained and frozen. The transfer objective is to preserve a
previously learned \emph{adversarial detection mechanism} while adapting the
intermediate representations supplied to that detector.

\subsection{Detector Generalization and Reuse}
\label{subsec:detector_generalization}

Several studies examine whether adversarial detectors remain effective when
the attack, dataset, or protected model changes. DAMAD evaluates detection
across databases, attacks, and models using fused texture and autoencoder
features \cite{9377649}. Other approaches improve
cross-attack generalization through adversarial domain adaptation or few-shot learning\cite{10.1145/3706061,liu2024enhancing}. \cite{LI2024103791} reduces dependence on internal classifier
representations by emphasizing high-frequency adversarial artifacts and
reconstructed edge information. Prediction-consistency
methods instead compare outputs from a protected model and an auxiliary model
to obtain a detection signal \cite{han2025prediction}.

These methods improve generalization by designing detection signals that are
less tied to a particular internal representation or by training detectors
for broader operating conditions. GraphRectify addresses a different problem.
We assume that a useful detector has already been learned around one source
backbone and ask whether its learned internal detection structure can be
\emph{reused} after the protected backbone changes.

This distinction matters in limited-data settings. Retraining a detector from
scratch on the target backbone discards the structure learned by the source
detector. Directly applying the source detector to the target backbone is also
unreliable because source and target intermediate features are not naturally
aligned. GraphRectify therefore treats backbone replacement as a
representation-compatibility problem: target features must be adapted before
the previously learned detector processes them.

\subsection{Graph-Based Internal Signal Modeling}
\label{subsec:graph_background}

Graph neural networks provide mechanisms for learning from relationships among
multiple representations. Graph convolutional networks aggregate neighboring
features according to graph connectivity \cite{kipf2016semi}, while graph
attention networks learn data-dependent weights over neighboring nodes
\cite{velivckovic2017graph}. More generally, graph representation learning
provides tools for modeling structured dependencies among multiple feature
sources \cite{hamilton2017representation}.

Graph-based techniques have also appeared in adversarial settings, although
with objectives different from ours. GNNGuard improves the robustness of graph
neural networks by reducing the influence of suspicious edges during message
passing \cite{zhang2020gnnguard}. NeuroTrace represents a neural network
forward pass as an input-dependent inference-provenance graph whose nodes and
edges encode intermediate activations and computational dependencies
\cite{hmida2026neurotrace}.

GraphRectify uses a graph for a different purpose. Its graph does not represent
an external relational dataset, and it does not reproduce the complete
computation graph of the classifier. Instead, selected intermediate feature
maps are compressed into a small set of detector nodes. The graph detector
learns relationships among these nodes that distinguish clean from adversarial
inputs.

This representation creates the transfer problem our method addresses. A
graph detector trained on source-backbone nodes expects not only individual
node features but also particular relationships among those features.
Consequently, matching dimensionality alone is insufficient when the backbone
changes. The target nodes must become compatible with the relational structure
learned by the source graph detector.

\subsection{Positioning of Our Work}
\label{subsec:positioning}

GraphRectify is motivated by a gap between detector generalization and detector
reuse. Existing work has studied detectors that generalize across attacks,
datasets, or models, and transfer-learning methods that move predictive
knowledge between networks. Neither setting directly addresses the case in
which a detector built from one classifier's intermediate representations must
be retained after a different backbone replaces that classifier.

Our approach separates the detector into two roles. The projection modules
provide a backbone-specific interface that converts intermediate features into
a common node representation, while the graph module captures a reusable
relational detection structure. When transferring to a target backbone, we
freeze the learned graph detector and adapt the target-side interface instead
of relearning the graph-level detector.

The key mechanism is source-guided bounded rectification. Target nodes are
paired with detached source reference nodes obtained from the same input, and a
lightweight rectifier predicts a bounded correction before graph processing.
The source branch therefore guides how the target representation is adapted,
but it does not provide the final detection prediction and does not transfer
source-classifier class outputs. This allows GraphRectify to preserve a learned
graph-level detection structure while adapting it to a new backbone using
limited target data.

\section{Proposed Method}
\label{sec:proposed_method}

GraphRectify transfers an adversarial detector from a source classifier
backbone to a different target backbone. The framework has three main
components: (1) backbone-specific projection modules that convert selected
intermediate feature maps into fixed-dimensional detector nodes, (2) a graph
module and readout network that model relationships among these nodes and
produce an adversarial-detection score, and (3) target-side bounded rectifiers
that adapt target nodes using detached reference nodes from the frozen source
branch.

The method proceeds in two stages. First, the source projection modules, graph
module, and readout network are trained jointly while the source classifier
remains frozen. Second, the learned graph module and readout are transferred to
the target path and frozen. The target projection modules and bounded
rectifiers are then optimized so that target representations become compatible
with the graph-level detection structure learned on the source backbone.

\subsection{Problem Setting}

Let $b\in\{s,t\}$ denote either the source branch ($b=s$) or the target branch
($b=t$), and let $f_b$ denote its frozen image-classification backbone. The
source and target backbones solve the same $C$-class classification task but
may have different architectures. For an input image $x$ with true class
label $y$, the output $f_b(x)\in\mathbb{R}^{C}$ contains the corresponding
class logits. The adversarial detector is a separate binary model in which
label $0$ denotes a clean input and label $1$ denotes an adversarial input.

Let $\mathcal{D}_{\mathrm{src}}$ denote the data used to train the source
detector. Let $\mathcal{D}_{\mathrm{adapt}}$ denote the subset of data from
the target-backbone setting used to adapt the detector to the target backbone.
Let $D_s$ denote the trained source detector, consisting of the source
projection modules, graph module, and readout network. Let $D_t$ denote the
adapted target detector, consisting of the target projection modules, bounded
rectifiers, and the graph module and readout transferred from $D_s$. The
classifier backbones remain separate from the detectors and are denoted
explicitly by $f_s$ and $f_t$.

For both branches, selected intermediate feature maps are converted into
compact nodes using global pooling and lightweight projection modules. We
refer to this as \emph{projection-only} node construction because no classifier
blocks or other architecture-specific modules are copied into the detector.
During target adaptation, the same input is evaluated by both frozen
backbones. The source branch provides detached reference nodes that guide
bounded corrections to the target nodes. Only the corrected target nodes are
used to produce the final target detection output.

\subsection{Base Classifiers and Adversarial Examples}

The source and target classifiers are trained independently on clean images
before detector learning and remain fixed during detector training and
adaptation. During source-detector training or target adaptation, adversarial
examples are generated against the corresponding frozen backbone.

Let $x_b^{\mathrm{adv}}$ denote an adversarial example generated against
$f_b$. Projected gradient descent starts from $x_b^{0}=x$ and updates
\begin{equation}
 x_b^{k+1}
 =
 \Pi_{\mathcal{B}_{\epsilon}(x)}
 \left(
 x_b^{k}
 +
 \delta
 \operatorname{sign}
 \left(
 \nabla_{x_b^{k}}
 \mathcal{L}_{\mathrm{ce}}
 \left(f_b(x_b^{k}),y\right)
 \right)
 \right),
 \label{eq:pgd}
\end{equation}
where $k$ is the attack-iteration index,
$\mathcal{B}_{\epsilon}(x)$ is the $\ell_{\infty}$ ball of radius
$\epsilon$ centered at $x$,
$\Pi_{\mathcal{B}_{\epsilon}(x)}$ denotes projection onto this set,
$\delta$ is the attack step size, and $\mathcal{L}_{\mathrm{ce}}$ is the
cross-entropy loss. This procedure is used to generate adversarial examples
during detector training and adaptation. The detector-aware attacks used for
final evaluation are described in Section~\ref{sec:experimental_setup}.

Let $x^{\mathrm{aug}}$ denote a weakly augmented version of the clean
input used by the contrastive objectives defined below. We construct
$x^{\mathrm{aug}}$ using the same weak training augmentation described
in Section~IV-A, including random cropping with four pixels of padding and random
horizontal flipping.

\subsection{Projection-Based Node Construction}

GraphRectify represents selected intermediate backbone features as a small set
of fixed-dimensional detector nodes. From each backbone, we extract feature
maps at $K$ selected intermediate stages, where each stage corresponds to a
network block or layer group.

Let $F_i^b(x)$ denote the feature map extracted from the $i$-th selected stage
of backbone $f_b$, where $i\in\{1,\ldots,K\}$. Let $P_i^b$ denote the
projection module associated with that stage. The corresponding detector node
is
\begin{equation}
 v_i^b(x)
 =
 P_i^b
 \left(
 \operatorname{pool}
 \left(F_i^b(x)\right)
 \right),
 \qquad
 v_i^b(x)\in\mathbb{R}^{d},
 \label{eq:detector_node}
\end{equation}
where $\operatorname{pool}(\cdot)$ denotes global pooling followed by
flattening, and $d$ is the common node dimension. Thus, projection modules
map intermediate features with potentially different channel dimensions into
a shared $d$-dimensional node space.

\begin{figure}[t]
\centering
\includegraphics[width=\linewidth]{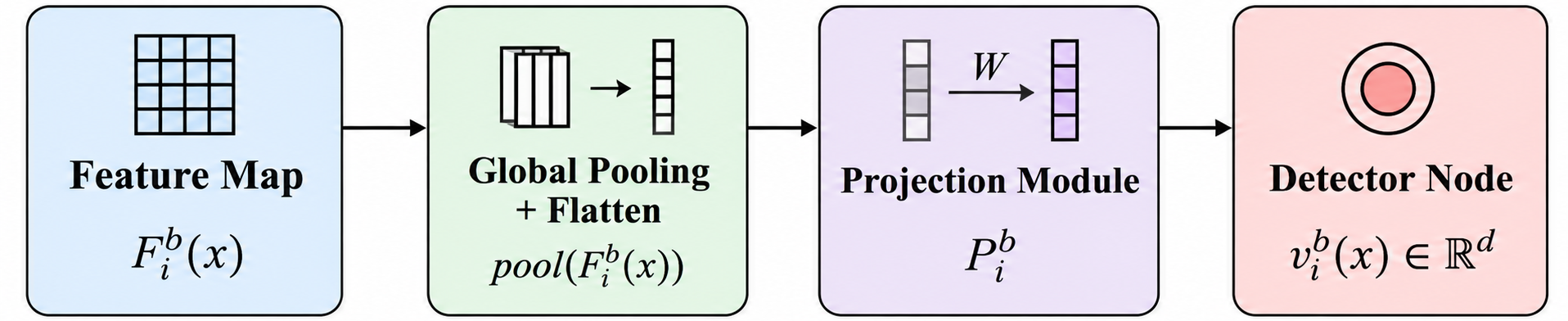}
\caption{Projection-based node construction. An intermediate feature map is
globally pooled and mapped by a lightweight projection module to a
$d$-dimensional detector node.}
\label{fig:node_construction}
\end{figure}

Figure~\ref{fig:node_construction} illustrates this operation.

\subsection{Graph-Based Detector}

The graph module receives a set of $K$ detector nodes
$\{v_i(x)\}_{i=1}^{K}$, each with dimension $d$. This notation is used
generically for the graph input. During source training, these are the source
nodes $v_i^s(x)$. After target adaptation is introduced below, the graph
receives corrected target nodes instead.

To construct relationships between intermediate representations, each node is
first normalized as
$\bar{v}_i(x)=v_i(x)/\lVert v_i(x)\rVert_2$. GraphRectify then constructs an
input-dependent adjacency matrix using the nonnegative cosine similarity
between each pair of nodes. A self-loop is added to every node, and each row
is normalized to sum to one. The resulting affinity from the $i$-th node to
the $j$-th node is
\begin{equation}
 A_{ij}(x)
 =
 \frac{
 \max\left(0,\bar{v}_i(x)^{\top}\bar{v}_j(x)\right)
 +
 \mathbb{I}[i=j]
 }{
 \sum_{n=1}^{K}
 \left(
 \max\left(0,\bar{v}_i(x)^{\top}\bar{v}_n(x)\right)
 +
 \mathbb{I}[i=n]
 \right)
 },
 \label{eq:adjacency}
\end{equation}
where $\mathbb{I}[\cdot]$ is the indicator function and $n$ indexes the nodes
used for row normalization. Because $A(x)$ is recomputed for each input, the
graph structure depends on the intermediate activation pattern produced by
that input.

Let $L$ denote the number of message-passing layers and
$h_i^{(\ell)}(x)$ the hidden state of the $i$-th node at the $\ell$-th graph
layer. The initial state is
\begin{equation}
 h_i^{(0)}(x)=v_i(x).
 \label{eq:graph_initialization}
\end{equation}
For $\ell\in\{0,\ldots,L-1\}$, message passing updates the node state as
\begin{equation}
 h_i^{(\ell+1)}(x)
 =
 \psi
 \left(
 W_{\mathrm{self}}^{(\ell)}h_i^{(\ell)}(x)
 +
 W_{\mathrm{neigh}}^{(\ell)}
 \sum_{j=1}^{K}
 A_{ij}(x)h_j^{(\ell)}(x)
 \right),
 \label{eq:message_passing}
\end{equation}
where $W_{\mathrm{self}}^{(\ell)}$ and
$W_{\mathrm{neigh}}^{(\ell)}$ are trainable weight matrices for the
$\ell$-th graph layer, and $\psi(\cdot)$ is its nonlinear activation.
The first term preserves information from the node itself, while the second
aggregates information from related nodes according to the dynamic adjacency
matrix.

After $L$ message-passing layers, GraphRectify summarizes the resulting node
states in several complementary ways. Let $g(\cdot)$ be a trainable scalar
scoring function. The importance weight of the $i$-th node is
\begin{equation}
 \omega_i(x)
 =
 \frac{
 \exp\left(g\left(h_i^{(L)}(x)\right)\right)
 }{
 \sum_{j=1}^{K}
 \exp\left(g\left(h_j^{(L)}(x)\right)\right)
 }.
 \label{eq:node_gate}
\end{equation}
The attention-weighted node summary is
\begin{equation}
 a(x)
 =
 \sum_{i=1}^{K}\omega_i(x)h_i^{(L)}(x),
 \label{eq:weighted_summary}
\end{equation}
while the element-wise max-pooled summary is
\begin{equation}
 m(x)
 =
 \operatorname*{max}_{i\in\{1,\ldots,K\}}
 h_i^{(L)}(x).
 \label{eq:max_summary}
\end{equation}

In addition to node summaries, the detector retains information about the
relationships among nodes. Let
$e(x)=\operatorname{vec}(A(x))$ denote the vectorized adjacency matrix, which
serves as an edge signature. We also define the graph energy as
\begin{equation}
 E_G(x)
 =
 \frac{1}{K}
 \sum_{i=1}^{K}
 \sum_{j=1}^{K}
 A_{ij}(x)
 \left\lVert
 h_i^{(L)}(x)-h_j^{(L)}(x)
 \right\rVert_2^2.
 \label{eq:graph_energy}
\end{equation}
The graph energy measures the weighted disagreement between connected final
node states.

The attention-weighted summary $a(x)$ emphasizes informative nodes,
$m(x)$ retains the strongest activation in each representation dimension,
$e(x)$ preserves the input-dependent connectivity pattern, and $E_G(x)$
summarizes disagreement among connected nodes. These components are
concatenated to form
\begin{equation}
 z(x)
 =
 \left[
 a(x),
 m(x),
 e(x),
 E_G(x)
 \right].
 \label{eq:graph_representation}
\end{equation}

Let $q(\cdot)$ denote the readout multilayer perceptron. When the graph receives
nodes from branch $b$, we denote the resulting graph representation, detector
logit, and adversarial-detection probability by
\begin{equation}
\begin{aligned}
z_b(x) &= z(x), \qquad
r_b(x) = q\left(z_b(x)\right), \\
p_b(x) &= \operatorname{sigmoid}\left(r_b(x)\right).
\end{aligned}
\label{eq:detector_output}
\end{equation}
Let $\mathcal{G}$ denote the complete graph module defined by
Eqs.~\eqref{eq:adjacency}--\eqref{eq:graph_representation}.

\begin{figure}[t]
\centering
\includegraphics[width=\linewidth]
{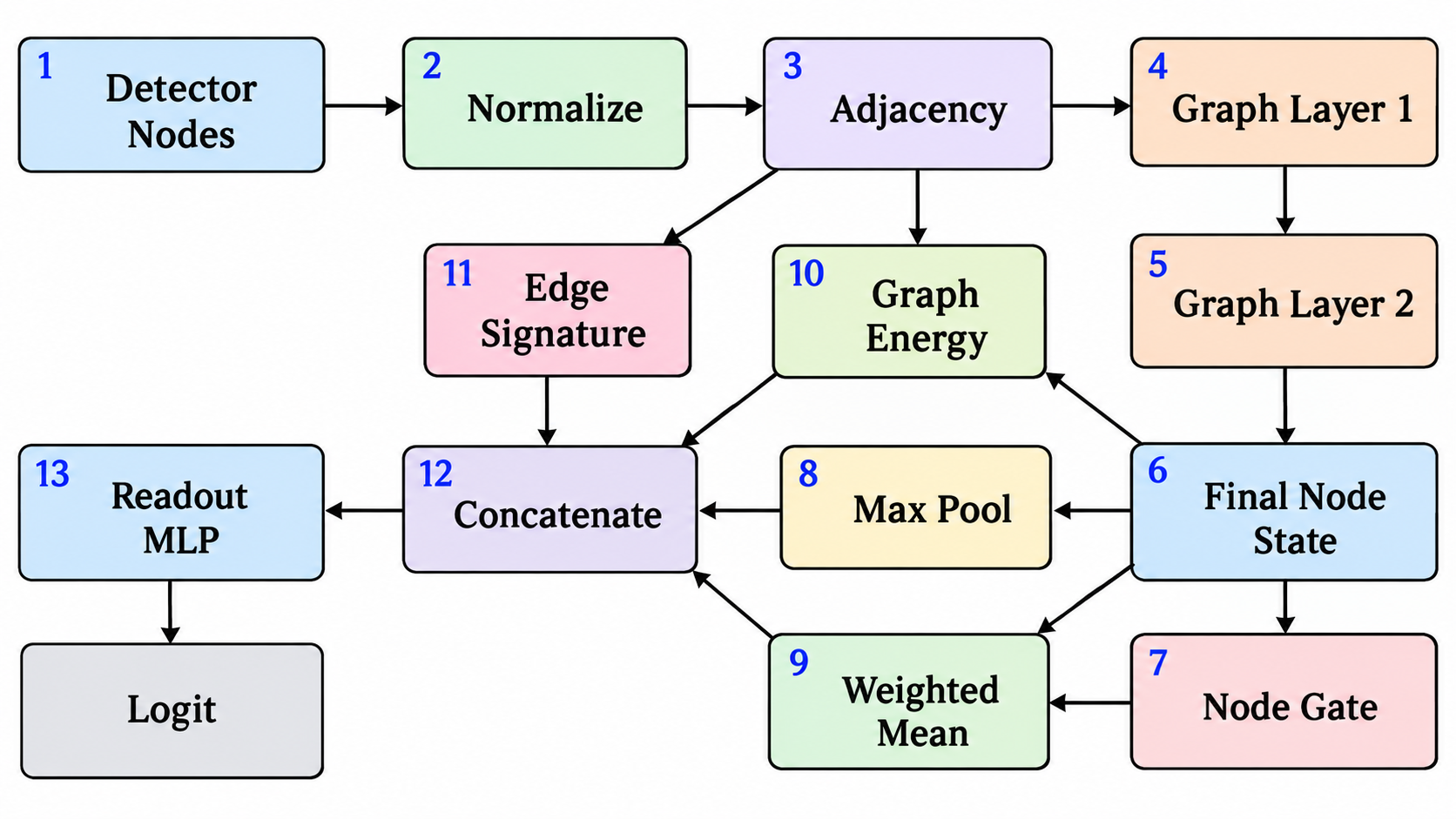}
\caption{Graph-based detector used in GraphRectify. Dynamic graph construction
and message passing are followed by node and edge aggregation and a readout
network that produces the adversarial-detection logit.}
\label{fig:gnn_head_architecture}
\end{figure}

Figure~\ref{fig:gnn_head_architecture} summarizes this process. The detector
first constructs the dynamic graph and applies message passing. It then
combines node-level summaries with the graph connectivity and energy before
the readout network produces the binary detection logit.

\subsection{Source Detector Training}

The first stage of GraphRectify learns the detection structure using the frozen
source backbone $f_s$. For each clean input $x$, PGD generates an adversarial
counterpart $x_s^{\mathrm{adv}}$ against $f_s$. The source projection modules
produce graph nodes for the clean, augmented, and adversarial inputs, and the
graph module and readout are trained jointly with these projections.

The binary detection term encourages the detector to assign label $0$ to the
clean input and label $1$ to its adversarial counterpart:
\begin{equation}
 \mathcal{L}_{\mathrm{det}}^{s}
 =
 \operatorname{BCEWithLogits}
 \left(
 \left[r_s(x),r_s(x_s^{\mathrm{adv}})\right],
 \left[0,1\right]
 \right).
 \label{eq:source_detection_loss}
\end{equation}

To preserve local consistency while separating adversarial representations,
we additionally use a contrastive term. Let $M$ denote the contrastive margin:
\begin{equation}
 \mathcal{L}_{\mathrm{con}}^{s}
 =
 \left\lVert z_s(x)-z_s(x^{\mathrm{aug}})\right\rVert_2^2
 +
 \max
 \left(
 0,
 M-
 \left\lVert z_s(x)-z_s(x_s^{\mathrm{adv}})\right\rVert_2
 \right)^2.
 \label{eq:source_contrastive_loss}
\end{equation}
The first term keeps the clean and weakly augmented graph representations
close, while the second encourages the adversarial representation to remain
at least margin $M$ away from the clean representation.

We also encourage separation in graph energy. Let $M_E$ denote the
graph-energy margin:
\begin{equation}
 \mathcal{L}_{\mathrm{eng}}^{s}
 =
 \max
 \left(
 0,
 M_E-
 \left(E_G(x_s^{\mathrm{adv}})-E_G(x)\right)
 \right).
 \label{eq:source_energy_loss}
\end{equation}
Here and below, $E_G(\cdot)$ is evaluated using the nodes supplied by the
active branch.

Let $\lambda_{\mathrm{con}}$ and $\lambda_E$ denote the weights of the
contrastive and graph-energy terms. The complete source objective is
\begin{equation}
\mathcal{L}_s
=
\mathcal{L}^{s}_{\mathrm{det}}
+
\lambda_{\mathrm{con}}\mathcal{L}^{s}_{\mathrm{con}}
+
\lambda_E\mathcal{L}^{s}_{\mathrm{eng}}.
\label{eq:source_objective}
\end{equation}

The source projection modules, $\mathcal{G}$, and $q$ are optimized using
$\mathcal{L}_s$, while $f_s$ remains fixed. The checkpoint with the highest
validation ROC-AUC is retained. These trained components form $D_s$ and are
then frozen for target adaptation. Algorithm~\ref{alg:source_training}
summarizes the procedure.

\begin{algorithm}[t]
\caption{Source detector training}
\label{alg:source_training}
\begin{algorithmic}[1]
\REQUIRE $f_s$, $\mathcal{D}_{\mathrm{src}}$,
$\{P_i^s\}_{i=1}^{K}$, $\mathcal{G}$, $q$, $\epsilon$, $\delta$
\ENSURE $D_s$

\STATE Freeze $f_s$
\FOR{each training epoch}
    \FOR{each mini-batch $(x,y)\sim\mathcal{D}_{\mathrm{src}}$}
        \STATE Generate $x_s^{\mathrm{adv}}$ using Eq.~\eqref{eq:pgd}
        \STATE Generate $x^{\mathrm{aug}}$
        \FOR{each $x'\in\{x,x^{\mathrm{aug}},x_s^{\mathrm{adv}}\}$}
            \STATE Extract $\{F_i^s(x')\}_{i=1}^{K}$ using $f_s$
            \STATE Compute $\{v_i^s(x')\}_{i=1}^{K}$ using
            Eq.~\eqref{eq:detector_node}
            \STATE Supply $\{v_i^s(x')\}_{i=1}^{K}$ to $\mathcal{G}$
            \STATE Compute $z_s(x')$, $r_s(x')$, and $E_G(x')$
        \ENDFOR
        \STATE Compute $\mathcal{L}_{\mathrm{det}}^{s}$,
        $\mathcal{L}_{\mathrm{con}}^{s}$, and
        $\mathcal{L}_{\mathrm{eng}}^{s}$
        \STATE Update $\{P_i^s\}_{i=1}^{K}$, $\mathcal{G}$, and $q$
        using Eq.~\eqref{eq:source_objective}
    \ENDFOR
    \STATE Evaluate validation ROC-AUC
\ENDFOR
\STATE Select the checkpoint with the highest validation ROC-AUC
\STATE Form and freeze $D_s$
\RETURN $D_s$
\end{algorithmic}
\end{algorithm}

\subsection{Transfer with Source-Guided Node Rectification}

After source training, the graph module $\mathcal{G}$ and readout $q$ learned
as part of $D_s$ are transferred to the target path and kept fixed. The target
projection modules $\{P_i^t\}_{i=1}^{K}$ are target-specific because the
intermediate feature dimensions of $f_t$ may differ from those of $f_s$.

The central transfer problem is that projecting target features into the same
$d$-dimensional space does not guarantee that they follow the representation
patterns expected by the graph detector learned on the source backbone.
GraphRectify therefore applies a bounded, source-guided correction to each
target node before graph reasoning.

For the $i$-th selected stage and input $x$, let $v_i^t(x)$ denote the target
node and $v_i^s(x)$ the corresponding detached source reference node. Let
$R_i(\cdot)$ denote the rectifier MLP for the $i$-th stage, $d_R$ its hidden
dimension, and $\eta$ the correction scale. The corrected target node is
\begin{equation}
 \tilde{v}_i^t(x)
 =
 v_i^t(x)
 +
 \eta
 \tanh
 \left(
 R_i
 \left(
 \left[v_i^t(x),v_i^s(x)\right]
 \right)
 \right).
 \label{eq:rectifier}
\end{equation}

Each rectifier maps the concatenated target and source nodes through two fully
connected layers with a nonlinear activation and outputs a residual vector in
$\mathbb{R}^{d}$. The hyperbolic tangent bounds each component of the learned
correction before scaling by $\eta$. Consequently, the target node remains the
base representation, and the rectifier can modify it only through a bounded
residual.

The source node is detached before being supplied to $R_i$, so gradients do
not update either the source backbone $f_s$ or the source detector $D_s$.
The source branch therefore provides reference information rather than a
prediction to be copied. The corrected nodes
$\{\tilde{v}_i^t(x)\}_{i=1}^{K}$ are supplied to the transferred
$\mathcal{G}$ and $q$ to obtain $z_t(x)$, $r_t(x)$, and $p_t(x)$. Thus, the
final detector output is determined by the corrected target path.

\subsection{Target Adaptation Objective}

Target adaptation trains only the target-specific interface to the frozen
source detector. PGD first generates $x_t^{\mathrm{adv}}$ against the frozen
target classifier $f_t$. For each
$x'\in\{x,x^{\mathrm{aug}},x_t^{\mathrm{adv}}\}$, the target and source
branches independently produce
$\{v_i^t(x')\}_{i=1}^{K}$ and $\{v_i^s(x')\}_{i=1}^{K}$. The source nodes are
detached and paired with the corresponding target nodes to obtain corrected
target nodes through Eq.~\eqref{eq:rectifier}. The adaptation path can be
written as
\begin{equation}
 x'
 \rightarrow
 \left(
 \{v_i^t(x')\}_{i=1}^{K},
 \{v_i^s(x')\}_{i=1}^{K}
 \right)
 \rightarrow
 \{\tilde{v}_i^t(x')\}_{i=1}^{K}
 \rightarrow
 \left(z_t(x'),r_t(x')\right).
 \label{eq:target_adaptation_path}
\end{equation}

The target detector uses the same three types of objectives as the source
detector. The binary detection loss is
\begin{equation}
 \mathcal{L}_{\mathrm{det}}^{t}
 =
 \operatorname{BCEWithLogits}
 \left(
 \left[r_t(x),r_t(x_t^{\mathrm{adv}})\right],
 \left[0,1\right]
 \right).
 \label{eq:target_detection_loss}
\end{equation}
The target contrastive loss is
\begin{equation}
 \mathcal{L}_{\mathrm{con}}^{t}
 =
 \left\lVert z_t(x)-z_t(x^{\mathrm{aug}})\right\rVert_2^2
 +
 \max
 \left(
 0,
 M-
 \left\lVert z_t(x)-z_t(x_t^{\mathrm{adv}})\right\rVert_2
 \right)^2,
 \label{eq:target_contrastive_loss}
\end{equation}
and the graph-energy loss is
\begin{equation}
 \mathcal{L}_{\mathrm{eng}}^{t}
 =
 \max
 \left(
 0,
 M_E-
 \left(E_G(x_t^{\mathrm{adv}})-E_G(x)\right)
 \right).
 \label{eq:target_energy_loss}
\end{equation}

The complete target adaptation objective is
\begin{equation}
\mathcal{L}_t
=
\mathcal{L}^{t}_{\mathrm{det}}
+
\lambda_{\mathrm{con}}\mathcal{L}^{t}_{\mathrm{con}}
+
\lambda_E\mathcal{L}^{t}_{\mathrm{eng}}.
\label{eq:target_objective}
\end{equation}

Only the target projection modules $\{P_i^t\}_{i=1}^{K}$ and rectifiers
$\{R_i\}_{i=1}^{K}$ are updated. Both classifier backbones, the source
detector $D_s$, and the transferred $\mathcal{G}$ and $q$ remain fixed.
Algorithm~\ref{alg:target_adaptation} summarizes the target-adaptation
procedure.

\begin{algorithm}[t]
\caption{GraphRectify target adaptation}
\label{alg:target_adaptation}
\begin{algorithmic}[1]
\REQUIRE $f_s$, $f_t$, $D_s$, $\mathcal{D}_{\mathrm{adapt}}$,
$\{P_i^t\}_{i=1}^{K}$, $\{R_i\}_{i=1}^{K}$, $\eta$
\ENSURE $D_t$

\STATE Freeze $f_s$, $f_t$, and $D_s$
\STATE Transfer and freeze $\mathcal{G}$ and $q$ from $D_s$
\FOR{each adaptation epoch}
    \FOR{each mini-batch $(x,y)\sim\mathcal{D}_{\mathrm{adapt}}$}
        \STATE Generate $x_t^{\mathrm{adv}}$ using Eq.~\eqref{eq:pgd}
        \STATE Generate $x^{\mathrm{aug}}$
        \FOR{each $x'\in\{x,x^{\mathrm{aug}},x_t^{\mathrm{adv}}\}$}
            \STATE Compute $\{v_i^t(x')\}_{i=1}^{K}$ using $f_t$
            and the target projections
            \STATE Compute detached $\{v_i^s(x')\}_{i=1}^{K}$
            using $f_s$ and $D_s$
            \STATE Compute $\{\tilde{v}_i^t(x')\}_{i=1}^{K}$
            using Eq.~\eqref{eq:rectifier}
            \STATE Supply $\{\tilde{v}_i^t(x')\}_{i=1}^{K}$ to
            the fixed $\mathcal{G}$ and $q$
            \STATE Compute $z_t(x')$, $r_t(x')$, and $E_G(x')$
        \ENDFOR
        \STATE Compute $\mathcal{L}_{\mathrm{det}}^{t}$,
        $\mathcal{L}_{\mathrm{con}}^{t}$, and
        $\mathcal{L}_{\mathrm{eng}}^{t}$
        \STATE Update $\{P_i^t\}_{i=1}^{K}$ and
        $\{R_i\}_{i=1}^{K}$ using Eq.~\eqref{eq:target_objective}
    \ENDFOR
    \STATE Evaluate validation ROC-AUC
\ENDFOR
\STATE Select the checkpoint with the highest validation ROC-AUC
\STATE Form and freeze $D_t$
\RETURN $D_t$
\end{algorithmic}
\end{algorithm}

\begin{figure*}[t]
\centering
\includegraphics[width=0.96\linewidth]
{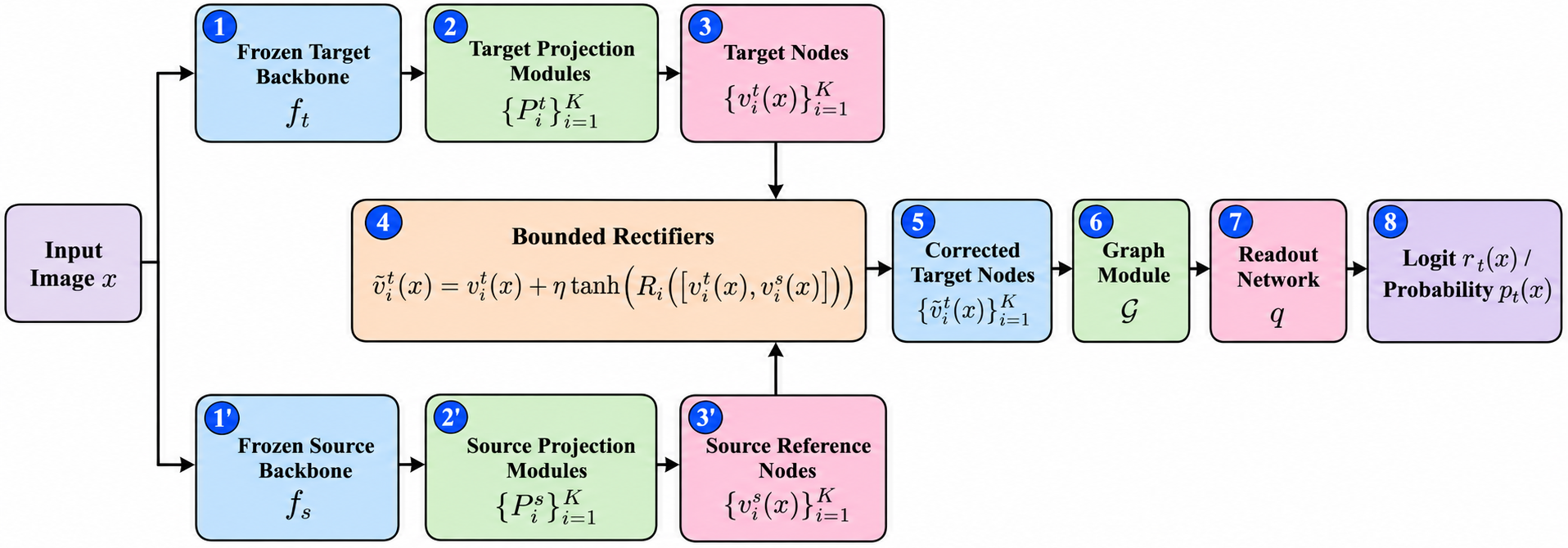}
\caption{GraphRectify target-transfer pipeline. Detached source reference nodes
guide bounded corrections to target nodes before the corrected target
representations are processed by the transferred graph module and readout.}
\label{fig:graphrectify_pipeline}
\end{figure*}

Figure~\ref{fig:graphrectify_pipeline} summarizes the complete target path.
For a given input, the frozen source and target backbones are evaluated in
parallel. Their selected intermediate features are transformed into detector
nodes by the corresponding projection modules. At each stage, the target node
is paired with its detached source reference node and corrected by the bounded
rectifier. Only the corrected target nodes are supplied to the transferred
graph module and readout. The source branch therefore guides representation
adaptation but does not directly produce the target detector output.

\subsection{Inference on the Target Backbone}

At inference, all components are fixed. Given a test image $x$, the target
backbone and its projection modules produce
$\{v_i^t(x)\}_{i=1}^{K}$, while the frozen source branch produces detached
reference nodes $\{v_i^s(x)\}_{i=1}^{K}$. Equation~\eqref{eq:rectifier} then
produces the corrected target nodes
$\{\tilde{v}_i^t(x)\}_{i=1}^{K}$.

These corrected target nodes are processed by the transferred graph module and
readout to obtain $z_t(x)$, $r_t(x)$, and the adversarial-detection
probability $p_t(x)$ from Eq.~\eqref{eq:detector_output}. The source branch is
used only to construct reference nodes. No loss is evaluated, and no parameters
are updated during inference.

\begin{algorithm}[t]
\caption{GraphRectify inference on the target backbone}
\label{alg:target_inference}
\begin{algorithmic}[1]
\REQUIRE $x$, $f_s$, $f_t$, $D_s$, $D_t$
\ENSURE $p_t(x)$

\STATE Compute $\{v_i^t(x)\}_{i=1}^{K}$ using $f_t$ and
the target projections in $D_t$
\STATE Compute detached $\{v_i^s(x)\}_{i=1}^{K}$ using $f_s$
and the source projections in $D_s$
\STATE Compute $\{\tilde{v}_i^t(x)\}_{i=1}^{K}$ using
Eq.~\eqref{eq:rectifier}
\STATE Supply $\{\tilde{v}_i^t(x)\}_{i=1}^{K}$ to the fixed
$\mathcal{G}$ and $q$
\STATE Compute $z_t(x)$ and $r_t(x)$
\STATE Compute $p_t(x)$ using Eq.~\eqref{eq:detector_output}
\RETURN $p_t(x)$
\end{algorithmic}
\end{algorithm}

\subsection{Computational Complexity}
\label{subsec:complexity}

Let $\Gamma_s$ and $\Gamma_t$ denote the costs of one forward pass through
$f_s$ and $f_t$, respectively. Constructing all pairwise node similarities
requires $O(K^2d)$ operations. Each message-passing layer also requires
$O(K^2d)$ operations, giving $O(LK^2d)$ across $L$ graph layers. We omit
lower-order projection and readout costs from the following expressions.

For the source path, the approximate cost per detector evaluation is
\begin{equation}
 O\left(
 \Gamma_s+(L+1)K^2d
 \right).
 \label{eq:source_complexity}
\end{equation}

The target path evaluates both frozen backbones. Each rectifier processes a
pair of $d$-dimensional nodes using hidden dimension $d_R$, giving
$O(Kdd_R)$ across the $K$ selected stages. The approximate source-guided
target cost is therefore
\begin{equation}
 O\left(
 \Gamma_t+\Gamma_s+Kdd_R+(L+1)K^2d
 \right).
 \label{eq:target_complexity}
\end{equation}

Thus, the graph-processing component scales primarily with $LK^2d$, while
source-guided transfer additionally requires the rectification term
$Kdd_R$ and a frozen forward pass through $f_s$. Compared with a target-only
detector, this source-backbone evaluation is the main additional inference
requirement. The source branch remains fixed and does not itself produce the
final target prediction.

\section{Experimental Setup}
\label{sec:experimental_setup}

This section describes the datasets, model architectures, transfer settings, training protocol, adaptive attacks, and evaluation metrics used to evaluate GraphRectify. To ensure complete, directly comparable comparisons, we evaluate every method and ablation using the same datasets, architecture transfers, target-data fractions, attacks, and random seeds.

\subsection{Datasets, Splits, and Hardware}

We evaluate GraphRectify on CIFAR-10, STL-10, and SVHN. These datasets provide complementary evaluation settings: CIFAR-10 contains low-resolution natural images, STL-10 contains higher-resolution natural images with substantially fewer labeled examples, and SVHN contains visually distinct street-number images.

For each dataset, we pool the available labeled images and construct a deterministic, disjoint split containing $70\%$ training examples, $10\%$ validation examples, and $20\%$ test examples. We use the training partition to train the image classifiers, source detectors, target-only detectors, and transferred detector components. The validation partition is used exclusively for checkpoint selection, and the test partition is reserved for final evaluation.

All backbones receive three-channel $32\times32$ inputs. STL-10 images are therefore downsampled from $96\times96$ to $32\times32$. Training augmentation consists of random cropping with four pixels of padding and, for CIFAR-10 and STL-10, random horizontal flipping. Horizontal flipping is disabled for SVHN because it can alter digit semantics. Inputs are normalized using dataset-specific channel statistics.

The experiments use a batch size of 128. Detector training is limited to at most 20 mini-batches per epoch, and detector validation uses at most 512 examples. Each final detector evaluation uses 1,024 test examples. We implemented the completed experiments using PyTorch 2.7 and torchvision 0.22 and ran them on a single NVIDIA H100 NVL GPU. Data loading used a single process to avoid accumulating worker processes and file descriptors over the large experimental matrix.

\subsection{Architectures and Transfer Matrix}

We consider three source-to-target architecture transfers:
\begin{equation}
\begin{aligned}
\mathrm{ResNet18} &\rightarrow \mathrm{ResNet34},
\qquad
\mathrm{ResNet34} \rightarrow \mathrm{ResNet18},\\
\mathrm{ResNet18} &\rightarrow \mathrm{MobileNetV2}.
\end{aligned}
\end{equation}
All three transfers are evaluated on CIFAR-10, STL-10, and SVHN.

For each backbone, we extract features from $K=4$ intermediate stages.
For ResNet18 and ResNet34, these stages correspond to the outputs of
\texttt{layer1}, \texttt{layer2}, \texttt{layer3}, and \texttt{layer4}.
For MobileNetV2, we use the outputs of feature layers with zero-based
indices 3, 6, 13, and 18 in the torchvision \texttt{features} sequence.
The selected stages are ordered from shallow to deep so that the $i$-th
source stage is paired with the $i$-th target stage during source-guided
rectification. All selected features are globally pooled and projected
to the common node dimension $d=128$.

We train the source and target image classifiers before detector learning and keep them frozen during source-detector training, target adaptation, and evaluation. CIFAR-10 classifiers are trained for 40 epochs. We train STL-10 ResNet18 and ResNet34 classifiers for 40 epochs, while we train STL-10 MobileNetV2 for 120 epochs because the shorter schedule produced insufficient classification accuracy. SVHN classifiers are trained for 80 epochs.

Classifier optimization uses stochastic gradient descent with an initial learning rate of $0.1$, momentum $0.9$, Nesterov acceleration, and weight decay $5\times10^{-4}$. We apply a cosine-annealing learning-rate schedule over the full training schedule. The checkpoint with the highest validation accuracy is retained. Before detector training, we audit each classifier using its
validation accuracy and reject a run if the validation accuracy
is below 0.5. The held-out test partition is not consulted during
classifier or detector selection and is used only for final evaluation.

All detector configurations use projection-only node construction. Intermediate backbone features are globally pooled and projected to a common node dimension $d=128$. The dynamic graph head contains $L=2$ message-passing layers with dropout $0.1$. Each feature rectifier has hidden dimension $d_R=64$ and correction scale $\eta=0.1$.

Each projection module $P_i^b$ consists of a fully connected layer that
maps the globally pooled stage features to the common node dimension
$d=128$, followed by a GELU activation and layer normalization. The
graph message-passing nonlinearity $\psi$ is GELU. The scalar scoring
function $g(\cdot)$ used for attention weighting is implemented as a
single fully connected layer from $\mathbb{R}^{d}$ to $\mathbb{R}$,
with no additional activation. The readout $q(\cdot)$ is a two-layer
multilayer perceptron that maps the concatenated graph representation
from dimension $2d+K^2+1$ to a hidden layer of dimension $d$ and then
to a scalar detection logit, with GELU between the two layers. Each
rectifier $R_i$ contains two fully connected layers with dimensions
$2d\rightarrow d_R\rightarrow d$, where $d_R=64$, and uses GELU
between the layers. Dropout with probability $0.1$ is applied after
the GELU activation in each graph message-passing layer and after the
GELU activation in the readout MLP. It is not applied in the projection
modules, attention-scoring layer, or rectifiers.

\subsection{Detector Training and Adaptation}

We train the source detector $D_s$ on the full source training partition and then freeze it. Source detectors are trained for 10 epochs on CIFAR-10 and STL-10 and for 30 epochs on SVHN. During transfer, we keep the source graph head and both image-classification backbones frozen. We optimize only the target projection layers and, when present, the rectifier parameters.

For every dataset and architecture transfer, we evaluate target-data fractions of 1\%, 5\%, 10\%, 25\%, and 100\%. We adapt CIFAR-10 and STL-10 for five epochs, while we adapt SVHN for 30 epochs. The selected target examples are deterministic for a given random seed.

Detector optimization uses AdamW with learning rate $5\times10^{-4}$ and weight decay $5\times10^{-4}$. Gradients are clipped to a maximum norm of $1.0$. The training objective uses contrastive weight
$\lambda_{\mathrm{con}} = 0.1$, contrastive margin $M = 1.0$,
effective graph-energy weight $\lambda_E = 10^{-4}$, and
graph-energy margin $M_E = 1.0$. The same objective weights are
used for both source-detector training and target adaptation. We validate the detector after every epoch and retain the checkpoint with the highest validation ROC-AUC.

We compare the complete GraphRectify model with four controlled baselines and ablations:
\begin{itemize}
    \item \textbf{Direct transfer:} target-backbone features are mapped to
the common $d$-dimensional node space using target-specific projection
modules. When the source and target stage dimensions match, these modules
are initialized from the corresponding source projections; otherwise,
they retain their seeded random initialization. The target projections
remain fixed, and the frozen source graph module and readout are applied
to the resulting target nodes without target adaptation or rectification;
    \item \textbf{Target-only:} the same projection, graph, and readout
architecture used by the source detector is newly instantiated on the
frozen target backbone. These detector components are trained from
scratch using only the corresponding target-data subset. No detector
parameters are transferred from the source branch, no source-reference
nodes are used, and no rectification modules are included. Target-only
training uses the same detector objective, optimizer, epoch budget, and
mini-batch limit as the corresponding transferred detector;
    \item \textbf{No rectifier:} the target projection modules are trained
on the target adaptation data, while the rectifiers are removed. The
transferred graph module and readout remain frozen;
    \item \textbf{No source reference}: the rectifiers are retained, but their source-reference inputs are replaced with zeros;
    \item \textbf{GraphRectify}: the complete transferred detector with target projections, source references, and learned feature rectification.
\end{itemize}

The direct-transfer baseline does not depend on the target-data fraction because it performs no target adaptation. Consequently, we evaluate it once for each dataset, transfer, seed, and attack, and reuse the result across fraction columns to maintain a rectangular comparison table without repeating the same computation.

\subsection{Attack Settings}

During detector training, we generate adversarial examples using PGD-10 with an $\ell_\infty$ perturbation budget of $\epsilon=8/255$, a step size of $2/255$, and random initialization. Checkpoint validation uses classifier-directed PGD-20 with the same perturbation budget and step size. We transform perturbations and projection bounds according to each dataset's normalization statistics so $\epsilon$ and the step size retain their pixel-space interpretations.

Final evaluation uses four detector-aware attacks for every dataset, architecture transfer, target-data fraction, method, and seed: FGSM, PGD-20, an APGD-style attack, and a CW-style projected $\ell_\infty$ attack. FGSM uses one step of size $8/255$ without random initialization. PGD-20 uses 20 steps of size $2/255$. The CW-style attack uses 40 projected steps and replaces cross-entropy with a logit-margin objective. The APGD-style attack uses 100 iterations, per-example best-solution tracking, a momentum update, and scheduled step-size reduction. We refer to this attack as APGD-style because it is an adaptive projected-gradient implementation rather than the canonical AutoAttack implementation.

Each attack jointly seeks to misclassify the target classifier and evade the detector. Let $\xi$ denote the perturbation. The adaptive objective is
\begin{equation}
\begin{aligned}
\max_{\lVert\xi\rVert_{\infty}\leq\epsilon}\quad
&\lambda_{\mathrm{cls}}
\mathcal{L}_{\mathrm{cls}}
\left(f_t(x+\xi),y\right) \\
&-
\lambda_{\mathrm{det}}
\operatorname{BCEWithLogits}
\left(r_t(x+\xi),0\right).
\end{aligned}
\label{eq:adaptive_attack_objective}
\end{equation}
where $\mathcal{L}_{\mathrm{cls}}$ is the cross-entropy loss for FGSM, PGD-20, and the APGD-style attack, and the CW logit-margin objective for the CW-style attack. The detector term encourages the adversarial example to receive the clean label $0$. We set $\lambda_{\mathrm{cls}}=\lambda_{\mathrm{det}}=1.0$ in every experiment.

For a fixed dataset, transfer, fraction, seed, and attack, all methods use the same evaluation examples and random attack initialization. This pairing ensures that differences between detector variants are not caused by attack randomness.

\subsection{Evaluation Protocol and Metrics}

The primary metric is ROC-AUC, which evaluates the detector's ranking quality regardless of the decision threshold. We additionally report detection accuracy using a threshold of $0.5$, clean target-classifier accuracy, and adversarial target-classifier accuracy. The classifier metrics verify that clean classification remains meaningful and that the generated adversarial examples successfully attack the frozen target classifier.

The complete evaluation matrix contains three datasets, three architecture transfers, five target-data fractions, five detector variants, four adaptive attacks, and three seeds ($11, 22, 33$). This produces 
$3\times3\times5\times5\times4\times3=2700$ seed-level evaluation cells. Every reported method and ablation is therefore evaluated under the same experimental coverage. Results and completion metadata are saved after each task, allowing interrupted runs to resume without repeating completed experiments, and all seed-level and aggregated results are consolidated into a single output file.

\section{Results}
\label{sec:results}

We evaluate all five detector variants using the same three datasets, three
architecture transfers, five target-data fractions, four detector-aware attacks,
and three random seeds. The complete matrix contains 2,700 seed-level result cells. All classifier checkpoints passed the quality audit across the three
datasets, three architectures, and three random seeds. Unless stated otherwise, each result is the mean $\pm$ sample
standard deviation over seeds 11, 22, and 33. ROC-AUC is the primary metric. We report detection accuracy separately because it depends on the fixed decision threshold.

\subsection{Performance across Datasets and Target-Data Fractions}

Table~\ref{tab:dataset_fraction_results} gives the main comparison. To avoid
selecting individual transfers or attacks, each seed-level entry is first
averaged over all three architecture transfers and all four attacks. The table
then reports the mean and standard deviation of those three seed-level averages.
The direct-transfer detector is independent of the target-data fraction, so its
single value is shown once for each dataset rather than repeated in every row.

Across the complete evaluation matrix, GraphRectify obtains an ROC-AUC of
$0.877\pm0.013$, compared with $0.840\pm0.016$ for target-only training. The
paired mean improvement is $0.037\pm0.011$ across the three seeds, and the
improvement is positive for every seed. GraphRectify also exceeds the no-rectifier and no-source-reference variants, which obtain $0.819\pm0.008$ and
$0.819\pm0.010$, respectively. Thus, the aggregate gain cannot be attributed
only to target-side detector training or to additional target-side parameters.

The amount of target data determines when transfer is beneficial. Averaged over
datasets, transfers, and attacks, GraphRectify increases from $0.764$ ROC-AUC at
1\% target data to $0.878$, $0.903$, $0.918$, and $0.918$ at 5\%, 10\%, 25\%,
and 100\%, respectively. At 1\%, target-only training is stronger on average
($0.789$ versus $0.764$). From 5\% onward, GraphRectify has the higher aggregate
ROC-AUC. A plausible explanation is that, with extremely few target examples,
fitting target projections and rectifiers is less reliable than training the
simpler target-only detector.

The dataset-level results further qualify the aggregate conclusion. On CIFAR-10,
GraphRectify reaches $0.915\pm0.021$ ROC-AUC compared with $0.897\pm0.017$ for
target-only training. On SVHN, the corresponding values are $0.938\pm0.005$ and
$0.819\pm0.021$. STL-10 is more difficult: GraphRectify is strongest at 25\%
and 100\% target data, while target-only training is stronger in the lower-data
regimes. This behavior is consistent with the smaller labeled dataset and the
lower STL-10 classifier accuracies, which leave less reliable target features for
estimating the transfer modules.

\begin{table*}[!t]
\caption{ROC-AUC across datasets and target-data fractions. For each seed, values
are averaged over all three architecture transfers and all four detector-aware
attacks; the table reports mean $\pm$ sample standard deviation over three seeds.
Bold indicates the highest mean in each row. Direct transfer does not use target
data and is therefore constant across fractions.}
\label{tab:dataset_fraction_results}
\centering
\begingroup
\normalsize
\setlength{\tabcolsep}{5.5pt}
\renewcommand{\arraystretch}{1.12}

\begin{tabular}{@{}llccccc@{}}
\toprule
Dataset & Target data & Direct & Target-only & No rectifier & No source ref. & GraphRectify \\
\midrule

\multirow{5}{*}{CIFAR-10}
 & 1\%   & \multirow{5}{*}{$0.336\pm0.122$} & $\mathbf{0.867\pm0.011}$ & $0.805\pm0.028$ & $0.805\pm0.028$ & $0.826\pm0.021$ \\
 & 5\%   & & $0.910\pm0.005$ & $0.897\pm0.030$ & $0.896\pm0.029$ & $\mathbf{0.933\pm0.030}$ \\
 & 10\%  & & $0.916\pm0.009$ & $0.898\pm0.019$ & $0.901\pm0.019$ & $\mathbf{0.943\pm0.019}$ \\
 & 25\%  & & $0.905\pm0.040$ & $0.899\pm0.016$ & $0.901\pm0.016$ & $\mathbf{0.942\pm0.017}$ \\
 & 100\% & & $0.888\pm0.021$ & $0.896\pm0.010$ & $0.893\pm0.013$ & $\mathbf{0.932\pm0.020}$ \\

\midrule

\multirow{5}{*}{STL-10}
 & 1\%   & \multirow{5}{*}{$0.400\pm0.032$} & $\mathbf{0.727\pm0.033}$ & $0.562\pm0.021$ & $0.564\pm0.026$ & $0.564\pm0.030$ \\
 & 5\%   & & $\mathbf{0.802\pm0.033}$ & $0.749\pm0.011$ & $0.753\pm0.019$ & $0.758\pm0.018$ \\
 & 10\%  & & $\mathbf{0.821\pm0.023}$ & $0.813\pm0.024$ & $0.807\pm0.019$ & $0.819\pm0.028$ \\
 & 25\%  & & $0.824\pm0.010$ & $0.847\pm0.013$ & $0.842\pm0.020$ & $\mathbf{0.869\pm0.024}$ \\
 & 100\% & & $0.840\pm0.031$ & $0.832\pm0.032$ & $0.838\pm0.033$ & $\mathbf{0.870\pm0.028}$ \\

\midrule

\multirow{5}{*}{SVHN}
 & 1\%   & \multirow{5}{*}{$0.161\pm0.085$} & $0.771\pm0.038$ & $0.763\pm0.026$ & $0.778\pm0.008$ & $\mathbf{0.903\pm0.022}$ \\
 & 5\%   & & $0.838\pm0.030$ & $0.841\pm0.016$ & $0.835\pm0.022$ & $\mathbf{0.944\pm0.003}$ \\
 & 10\%  & & $0.830\pm0.026$ & $0.822\pm0.012$ & $0.823\pm0.012$ & $\mathbf{0.947\pm0.014}$ \\
 & 25\%  & & $0.834\pm0.019$ & $0.830\pm0.023$ & $0.825\pm0.011$ & $\mathbf{0.943\pm0.025}$ \\
 & 100\% & & $0.821\pm0.033$ & $0.826\pm0.031$ & $0.817\pm0.007$ & $\mathbf{0.954\pm0.006}$ \\

\bottomrule
\end{tabular}

\endgroup
\end{table*}

\subsection{Robustness across Attacks and Architecture Transfers}

Table~\ref{tab:robustness_summary} separates the results by attack and transfer
without restricting the comparison to a selected dataset or target-data
fraction. GraphRectify gives the highest ROC-AUC under PGD-20, the APGD-style
attack, and the CW-style attack. Under FGSM, GraphRectify and target-only
training are effectively tied after rounding ($0.836$ for both). Thus, the
aggregate improvement is supported by the iterative attacks and is not driven by
a single evaluation objective.

GraphRectify also gives the highest aggregate ROC-AUC for all three architecture
transfers. The largest improvement occurs for the cross-family
ResNet18-to-MobileNetV2 transfer, where GraphRectify reaches
$0.850\pm0.014$, compared with $0.781\pm0.005$ for target-only training and
approximately $0.72$ for the two ablations. This result provides evidence that
the transferred graph structure remains useful when the target backbone belongs
to a different architecture family.

The direct-transfer results are substantially lower than the adapted variants.
In particular, an ROC-AUC below $0.5$ means that the score ordering learned on
the source representation is not preserved on the target representation; it
does not indicate that the adapted experiments used a different evaluation
protocol. Because direct transfer is especially sensitive to representation
misalignment, our main efficacy claims rely on the paired target-only and
ablation comparisons rather than on direct transfer alone.

At the fixed threshold, GraphRectify obtains detection accuracy
$0.708\pm0.012$, compared with $0.665\pm0.013$ for target-only training. The
clean target-classifier accuracy is identical across detector variants
($0.850\pm0.005$) because the target classifiers and evaluation examples are
shared. Under attacks optimized against GraphRectify, adversarial classifier
accuracy falls to $0.0902\pm0.0003$, confirming that ineffective attacks do not produce the reported detection results.

\begin{table*}[!t]
\caption{Attack-wise, transfer-wise, and overall results. Attack rows average
over datasets, transfers, and target-data fractions; transfer rows average over
datasets, attacks, and fractions. Overall rows average over the complete matrix.
Each value is the mean $\pm$ sample standard deviation over three seed-level
averages. Bold marks the strongest detector result; lower adversarial classifier
accuracy indicates a stronger successful attack.}
\label{tab:robustness_summary}
\centering
\begingroup
\normalsize
\setlength{\tabcolsep}{4pt}
\renewcommand{\arraystretch}{1.12}

\begin{tabular}{@{}lccccc@{}}
\toprule
Condition & Direct & Target-only & No rectifier & No source ref. & GraphRectify \\
\midrule

\multicolumn{6}{@{}l}{\textit{Attack-wise ROC-AUC}} \\
FGSM       & $0.264\pm0.026$ & $\mathbf{0.836\pm0.031}$ & $0.769\pm0.016$ & $0.774\pm0.016$ & $\mathbf{0.836\pm0.023}$ \\
PGD-20     & $0.287\pm0.050$ & $0.810\pm0.018$ & $0.811\pm0.006$ & $0.809\pm0.009$ & $\mathbf{0.874\pm0.011}$ \\
APGD-style & $0.319\pm0.044$ & $0.877\pm0.014$ & $0.869\pm0.009$ & $0.869\pm0.011$ & $\mathbf{0.912\pm0.007}$ \\
CW-style   & $0.326\pm0.040$ & $0.835\pm0.013$ & $0.825\pm0.007$ & $0.823\pm0.008$ & $\mathbf{0.885\pm0.010}$ \\

\midrule
\multicolumn{6}{@{}l}{\textit{Transfer-wise ROC-AUC}} \\
ResNet18 $\rightarrow$ ResNet34
& $0.444\pm0.134$ & $0.855\pm0.031$ & $0.853\pm0.013$ & $0.851\pm0.021$ & $\mathbf{0.879\pm0.019}$ \\
ResNet34 $\rightarrow$ ResNet18
& $0.311\pm0.085$ & $0.883\pm0.020$ & $0.885\pm0.015$ & $0.884\pm0.016$ & $\mathbf{0.900\pm0.018}$ \\
ResNet18 $\rightarrow$ MobileNetV2
& $0.141\pm0.058$ & $0.781\pm0.005$ & $0.717\pm0.020$ & $0.720\pm0.024$ & $\mathbf{0.850\pm0.014}$ \\

\midrule
\multicolumn{6}{@{}l}{\textit{Overall metrics}} \\
ROC-AUC
& $0.299\pm0.033$ & $0.840\pm0.016$ & $0.819\pm0.008$ & $0.819\pm0.010$ & $\mathbf{0.877\pm0.013}$ \\
Detection accuracy
& $0.425\pm0.020$ & $0.665\pm0.013$ & $0.664\pm0.009$ & $0.665\pm0.008$ & $\mathbf{0.708\pm0.012}$ \\
Adversarial classifier acc.
& $0.141\pm0.010$ & $0.100\pm0.004$ & $0.098\pm0.001$ & $0.098\pm0.001$ & $\mathbf{0.0902\pm0.0003}$ \\

\bottomrule
\end{tabular}

\endgroup
\end{table*}

\subsection{Ablation Evidence and Result Stability}

The ablations are evaluated over the same 540 seed-level conditions as the full
method. Relative to the no rectifier variant, GraphRectify improves mean ROC-AUC
by $0.058\pm0.006$ across seeds. Relative to the no source reference variant,
the improvement is $0.058\pm0.004$. The gain is positive for every seed in both
comparisons. Across the 180 dataset-transfer-fraction-attack conditions obtained
after averaging over seeds, GraphRectify is the highest-scoring method in 108
conditions and exceeds target-only training in 127 conditions. The remaining
conditions are concentrated in STL-10, the 1\% target-data regime, and FGSM,
which are the settings where the full transfer module has the smallest advantage.

Based on these results, we can conclude that GraphRectify provides a
consistent aggregate improvement, benefits from both bounded rectification and
source-reference guidance, and is particularly effective for SVHN and
cross-family transfer. Its advantage is strongest once sufficient target data are
available, and we should not interpret it as uniformly superior in the most
data-limited STL-10 settings.

\section{Discussion and Future Directions}
\label{sec:discussion}

The results show that detector transfer across classifier backbones requires
adaptation rather than direct reuse. Direct transfer performs substantially
worse than the adapted variants, indicating that detection behavior learned
from source-backbone representations is not directly preserved when the
backbone changes. GraphRectify addresses this mismatch by retaining the
source graph-level detection structure while adapting target representations
through projection and source-guided bounded rectification. Across the complete
evaluation matrix, GraphRectify achieves a mean ROC-AUC of $0.877$, compared
with $0.840$ for target-only training, and the improvement is positive for all
three random seeds.

The benefit of transfer depends on the amount and quality of target data.
GraphRectify is not uniformly superior in the most data-limited regime. With
only $1\%$ target data, target-only training performs better on average
($0.789$ versus $0.764$ ROC-AUC), whereas GraphRectify has the higher
aggregate ROC-AUC from $5\%$ target data onward. A similar limitation appears
on STL-10, where target-only training is stronger in several lower-data
settings. This suggests that the target projection and rectification modules
require sufficient target examples to estimate a useful mapping into the
representation space expected by the transferred graph detector. Future work
could therefore investigate more data-efficient adaptation, including
regularization, parameter sharing across stages, or initialization strategies
that reduce the amount of target data required.

The architecture-transfer results provide additional evidence for the value of
reusing the learned graph structure. GraphRectify improves over target-only
training for all three evaluated transfers, with its largest aggregate
advantage occurring for the cross-family
ResNet18$\rightarrow$MobileNetV2 setting. This suggests that the graph-level
detection structure can remain useful across different convolutional
architecture families when the target nodes are appropriately adapted.
However, the current experiments are limited to convolutional backbones.
An important next step is therefore to evaluate transfer involving
substantially different architectures, particularly Vision Transformers and
other token-based vision models. Such extensions require suitable definitions
of detector nodes from transformer representations, such as selected block
outputs, class-token embeddings, or pooled patch-token representations.

The ablation results further clarify the roles of the proposed components.
GraphRectify improves mean ROC-AUC by approximately $0.058$ over both the
no-rectifier and no-source-reference variants, with positive improvements for
every evaluated seed. These results indicate that the gain is not explained
only by additional trainable target-side parameters. Instead, both bounded
node correction and source-reference guidance contribute to transfer. Future
work could examine whether the source-reference branch can be compressed,
cached, or approximated while retaining this benefit.

The detector-aware evaluation also places the results in the appropriate
security context. GraphRectify performs better than target-only training under
the iterative PGD-20, APGD-style, and CW-style attacks, while the two methods
are approximately tied under FGSM. The adversarial classifier accuracy remains
low under attacks optimized jointly against the classifier and detector,
showing that the detection results are not explained by attacks that fail to
fool the target classifier. Nevertheless, these experiments cover a specific
$\ell_{\infty}$ threat model and a fixed set of adaptive attacks. Future work
should consider additional perturbation models, stronger adaptive strategies,
and settings in which new attacks arrive over time. In the latter case, the
lightweight target projection and rectification modules could be updated
incrementally while the transferred graph detector remains fixed.

Finally, GraphRectify introduces additional inference cost because both the
target backbone and the frozen source branch are evaluated to construct the
corrected target nodes. This cost is the main practical trade-off relative to
a target-only detector. Future work should therefore reduce reliance on the full source backbone during inference. This could be achieved by training a smaller reference encoder to reproduce the required source nodes or by distilling the source-guided corrections into the target branch, allowing the adapted detector to operate using only the target backbone at deployment.

Overall, the experiments support a qualified conclusion: reusing graph-level
detection structure together with source-guided target adaptation provides a
consistent aggregate improvement over target-only training and the two
transfer ablations, particularly when sufficient target data are available.
At the same time, the results identify clear directions for improving
data efficiency, architectural generality, adaptive robustness, and inference
cost.

\section{Conclusion}
\label{sec:conclusion}

We presented GraphRectify, a graph-based framework for transferring adversarial
example detectors across classifier backbones by reusing a learned graph-level
detection structure and adapting target representations through projection and
source-guided bounded rectification. Across three datasets, three architecture
transfers, five target-data fractions, and four detector-aware attacks,
GraphRectify achieves higher aggregate ROC-AUC than target-only training and
both transfer ablations, while the ablation results support source-conditioned rectification as a joint mechanism. The strongest gains appear once
a moderate amount of target data is available and in the cross-family
ResNet18$\rightarrow$MobileNetV2 setting, although target-only training can be
competitive or stronger in extremely data-limited settings. Overall, the
results show that previously learned adversarial detection structure can be
reused across changing classifier backbones rather than being relearned from
scratch for every new model.

\bibliographystyle{IEEEtran}
\bibliography{references}

\end{document}